\documentclass[letterpaper, 10 pt, conference]{ieeeconf}  
\usepackage{amsmath}
\usepackage{xcolor}
\usepackage{graphicx}
\usepackage{multirow}
\usepackage{wasysym}
\usepackage{amssymb}
\usepackage{float}
\usepackage[final]{hyperref}
\hypersetup{
  colorlinks=true,
  linkcolor=blue,
  urlcolor=cyan,
}
\IEEEoverridecommandlockouts                              

\title{\LARGE \bf
Enhancing Autonomous Vehicle Training with Language Model Integration and Critical Scenario Generation
}

\author{Hanlin Tian$^{1}$, Kethan Reddy$^{1}$, Yuxiang Feng$^{1}$, \\ Mohammed Quddus$^{1}$, Yiannis Demiris$^{2}$, and Panagiotis Angeloudis$^{1}$
\thanks{$^{1}$H. Tian, K. Reddy, Y. Feng, M. Quddus, and P. Angeloudis are with the Centre for Transport Engineering and Modelling, Department of Civil and Environmental Engineering, Imperial College London, UK
        {\tt\small h.tian22@imperial.ac.uk}}
\thanks{$^{2}$Y. Demiris is with the Personal Robotics Laboratory, Department of Electrical and Electronic Engineering, Imperial College London, UK
}
}

\begin{document}

\maketitle
\thispagestyle{empty}
\pagestyle{empty}

\begin{abstract}

This paper introduces CRITICAL, a novel closed-loop framework for autonomous vehicle (AV) training and testing. CRITICAL stands out for its ability to generate diverse scenarios, focusing on critical driving situations that target specific learning and performance gaps identified in the Reinforcement Learning (RL) agent. The framework achieves this by integrating real-world traffic dynamics, driving behavior analysis, surrogate safety measures, and an optional Large Language Model (LLM) component. It is proven that the establishment of a closed feedback loop between the data generation pipeline and the training process can enhance the learning rate during training, elevate overall system performance, and augment safety resilience. Our evaluations, conducted using the Proximal Policy Optimization (PPO) and the HighwayEnv simulation environment, demonstrate noticeable performance improvements with the integration of critical case generation and LLM analysis, indicating CRITICAL's potential to improve the robustness of AV systems and streamline the generation of critical scenarios. This ultimately serves to hasten the development of AV agents, expand the general scope of RL training, and ameliorate validation efforts for AV safety.
We make our code publicly available at \url{https://github.com/zachtian/CRITICAL}

\end{abstract}

\section{INTRODUCTION}
In the past decade, autonomous vehicles (AV) have achieved remarkable progress. This swift advancement can be attributed to improvements in the effectiveness of the collection of models that characterize an AV, such as perception, planning, localization, etc. Additionally, the implementation of robust risk metrics and safety assurances for handling critical situations, combined with the integration of these elements into a comprehensive control system that facilitates communication between modules, has played a crucial role \cite{ding2023survey}. A central component within the AI framework of an AV is the path planning and trajectory forecast module. The abundance of high-fidelity simulation environments and real-world scenario data collected over the years have allowed this decision-making process to be advantageously outsourced to reinforcement learning (RL) algorithms. The adaptive learning capability of RL is particularly beneficial in creating algorithms that can adeptly navigate dynamic and unpredictable road traffic environments \cite{DRL_survey}.
\begin{figure}[tp]
\centering
\includegraphics[width=.4\textwidth]{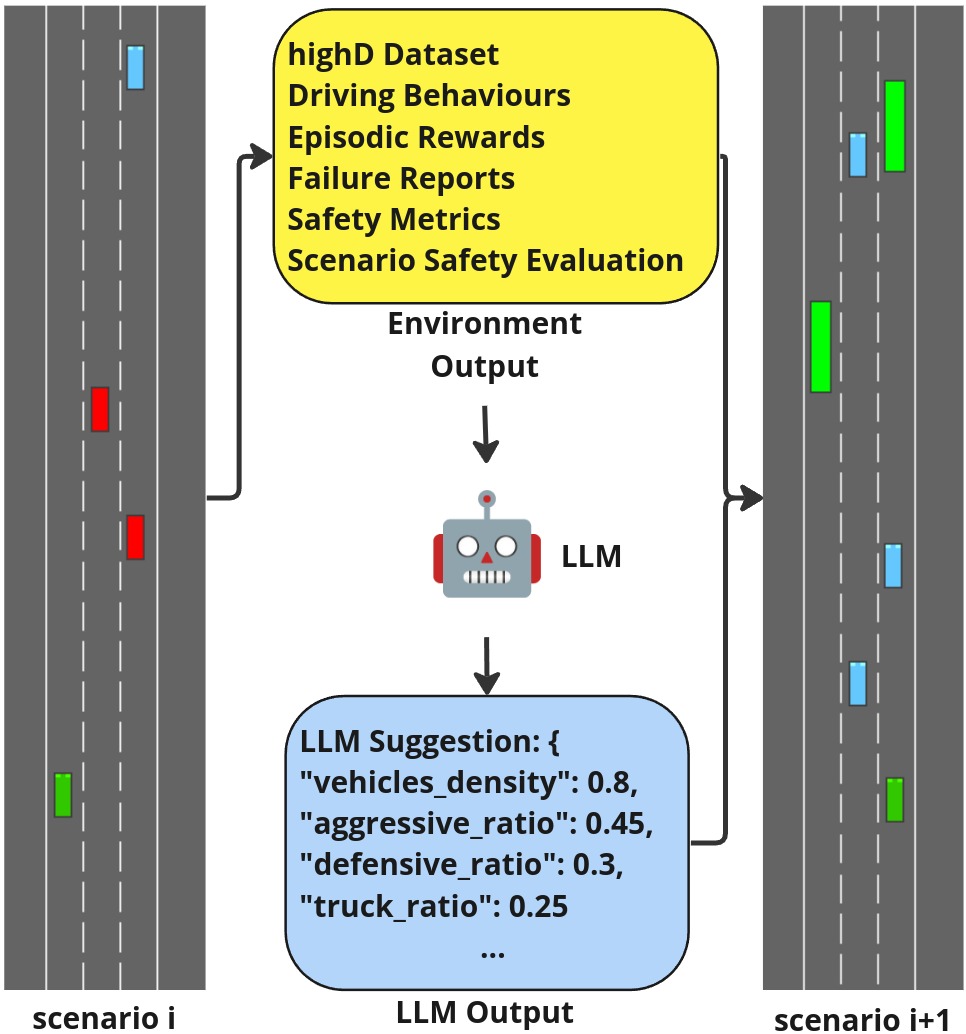}
\caption{A illustration of the general algorithmic flow of CRITICAL. An RL agent is exposed to a scenario (which is a function of the environment configuration), and after a designated number of episodes, we collate pertinent environment outputs (\textbf{yellow}) from every episode. This can then be used to directly generate a new scenario, or first be parsed into a prompt and fed into an LLM, to suggest an alternative environment configuration (\textbf{blue}) for instantiating scenarios.}
\label{fig:banner}
\end{figure}
However, the progress of open-source planning and prediction modules for RL in autonomous driving is impeded by the absence of an interpretable framework capable of generating diverse and pertinent safety-critical scenarios \cite{hao2023bridging}. Traditional AV training methods and environments often involve controlled, repeatable scenarios, which, while beneficial for initial learning, may not adequately cover the spectrum of real-world complexities \cite{krajewski}. And consequently, because the effectiveness of training RL for autonomous driving depends heavily on the variety and complexity of training scenarios provided, RL models exclusively trained in such environments might find themselves inadequately prepared to handle the unpredictable dynamics of road traffic \cite{Deo_2018}. A major challenge lies in accurately replicating the diverse conditions of real-world driving and addressing specific failure modes that AVs might encounter \cite{bolte2019corner}.

An intuitive method to bolster RL-based planning algorithms is to incorporate edge-case scenarios within an ego-vehicle's operational design domain (ODD) into the training process. A common approach is to use unsupervised clustering techniques, such as K-Means and Hierarchical Clustering Algorithms (HCA), to identify these ODD edge cases \cite{ponn2020identification}. These techniques, if used in isolation, fall short since they lack ground truth for validation. Other methodologies opt to utilize graph-based approaches to extract ontological relations between objects to sieve and generate a subset of edge-case scenarios from a set of boundary scenarios. While these techniques are effective, they are limited by their current intractable interpretability \cite{zipfl2023comprehensive}.

In this paper, we present a novel framework for training AVs that automatically generates critical scenarios, thereby augmenting conventional RL training. Additionally, we explore integrating Large Language Models (LLMs) to further refine these scenarios, leveraging real-world driving data to enrich training with diverse and challenging situations. By dissecting training episodes with LLMs to detect failure patterns, we significantly widen the scope of potential driving situations, enhancing the overall training process. This approach establishes a dynamic mechanism for the continuous, closed-loop refinement and validation of AV planning and prediction algorithms: we refer to this framework as "CRITICAL", and is overviewed at a high-level in Fig. \ref{fig:banner}.

The proposed framework is designed to isolate critical scenarios and leverage this identification to generate similar situations. This methodology combines data-driven and knowledge-driven approaches, leveraging the advantages of each while mitigating the drawbacks that arise when relying exclusively on either one \cite{hao2023bridging}. 

Our contributions are as follows:
\begin{enumerate}
\item We introduce a novel RL-based framework, CRITICAL, for enhancing the training and evaluation of AV algorithms. CRITICAL augments RL agents' exposure to a variety of scenarios, with a specific focus on critical driving scenarios to bolster AV performance and resilience.
\item Utilizing the highD dataset, we enrich our simulation environment with real-world traffic dynamics. Through clustering techniques, we analyze and replicate diverse driving behaviors, leveraging risk metrics to craft and incorporate high-fidelity critical scenarios into our training regimen. 
\item Our empirical findings validate that CRITICAL's closed-loop feedback mechanism between scenario generation and RL training significantly elevates the learning rate, overall AV performance, and adaptability to safety-critical situations.
\end{enumerate}

\section{RELATED WORK}

\subsection{Scenario Generation for Autonomous Vehicle}
In AV development, it is crucial to create a variety of challenging driving scenarios for evaluation purposes \cite{ding2023survey}. Data-driven generation methods in autonomous vehicle testing mainly rely on real-world traffic data to create diverse driving scenarios. Techniques range from direct sampling and clustering using Traffic Primitives\cite{wang2018extracting} to more complex strategies involving random perturbation for scenario augmentation \cite{fang2020augmented}. Additionally, advanced methods like Bayesian Networks for probabilistic modeling\cite{wheeler2016factor} and Deep Generative Models \cite{ding2018new} are utilized to generate realistic unseen traffic scenarios. 

Adversarial generation in autonomous vehicle testing actively creates high-risk scenarios by simulating challenging interactions. This approach involves a generator and a victim model \cite{david2021trustworthy}. For static scenarios, high-dimensional data generation is common, with techniques ranging from differentiable rendering for attacking detection algorithms \cite{jain2019analyzing} to optimization methods for manipulating point clouds and images \cite{abdelfattah2021towards}. Dynamic scenario generation is crucial for evaluating planning and control modules.

Knowledge-based generation in AV testing combines domain expertise with predefined rules to create realistic scenarios adhering to traffic laws and physical principles. Techniques range from optimizing dynamic object behaviors and environmental conditions \cite{o2018scalable} to employing RL for adversarial policy development \cite{chen2021adversarial}.

 \begin{figure}[t]
\centering
\includegraphics[width=.4\textwidth]{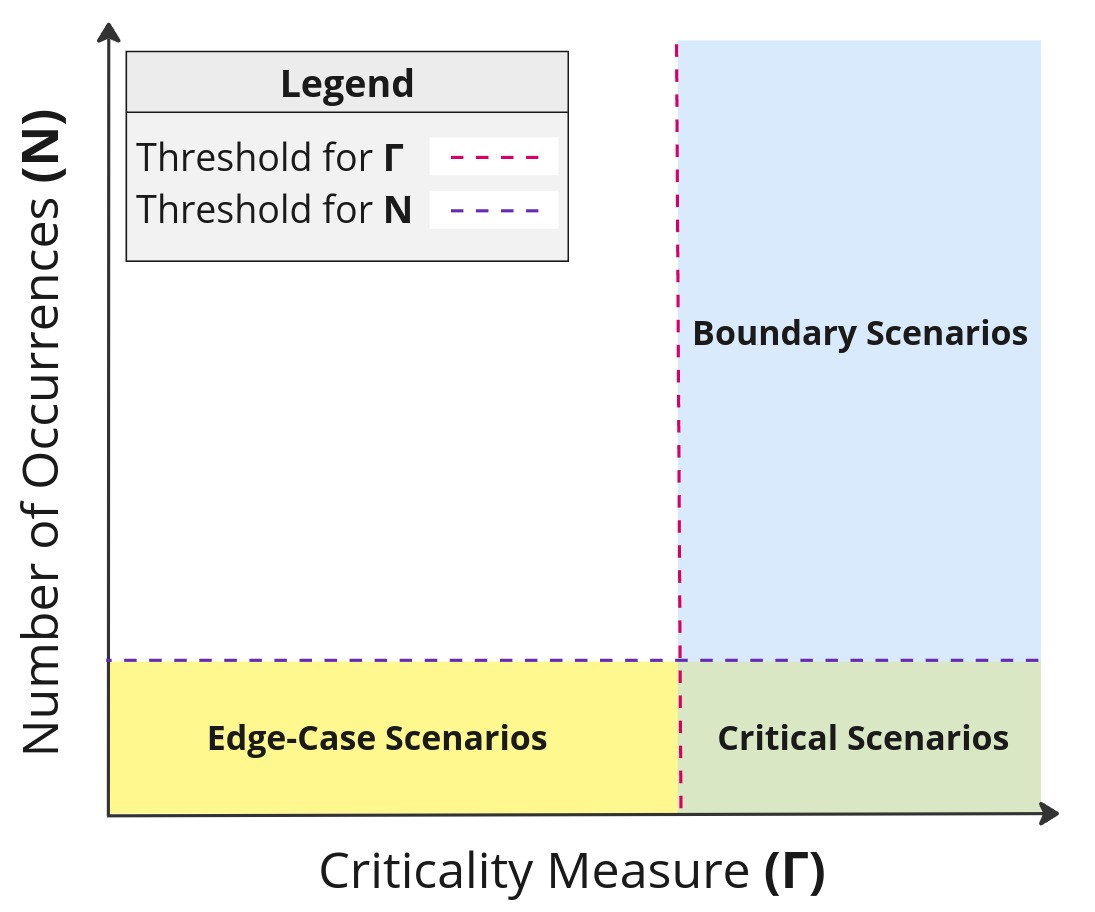}
\caption{Edge-case scenarios (\textbf{yellow}) are defined by their rarity in a given ODD. We depict edge-case scenarios to be below a certain threshold value of the number of occurrences \textit{N}. Boundary scenarios (\textbf{blue}) have criticality measure $\Gamma$ beyond a threshold value. Critical scenarios (\textbf{green}) are defined as scenarios that have the union of these two circumstances.}
\label{fig:Scenario}
\end{figure}

\begin{figure*}[t]
\centering
\includegraphics[width=\textwidth]{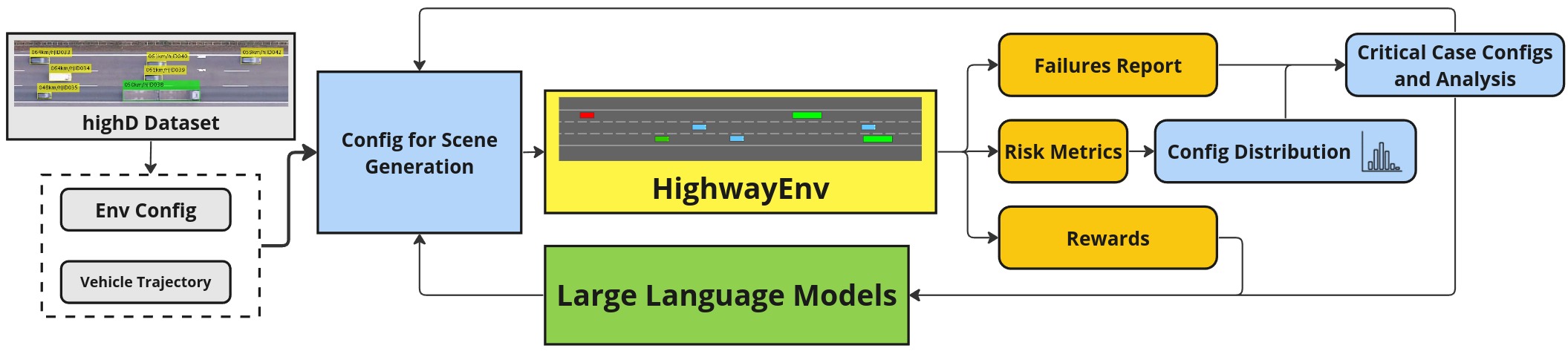}
\caption{A architecture diagram mapping out the various components of CRITICAL. The framework first sets up an environment configuration based on typical real-world traffic from the highD dataset \cite{highDdataset}. These configurations are then leveraged to generate HighwayEnv \cite{highway-env} scenarios. At the end of each episode, we collect data including failure reports, risk metrics, and rewards, repeating this process multiple times to gather a collection of configuration files with associated scenario risk assessments. To enhance RL training, we analyze a distribution of configurations based on risk metrics, identifying those conducive to critical scenarios. We then either directly use these configurations for new scenarios or prompt an LLM to generate critical scenarios.}
\label{fig:arch}
\end{figure*}

\subsection{Large Language Models in Autonomous Vehicles}\label{sec:LLMs}

LLMs have significantly evolved in recent years, exhibiting profound capabilities in text generation, comprehension, and other areas \cite{zhao2023survey}. The recent unveiling of Llama 2 distinctly marks a new phase; epitomizing collaborative advancements within the AI community through its open-source community license \cite{touvron2023llama}.

Furthermore, LLMs have showcased remarkable potential in reasoning-intensive question-answering scenarios \cite{madaan2022language}. The rising proclivity of prompt tuning over traditional fine-tuning techniques (in certain domains) accentuates the growing importance of the former for tailored tasks \cite{lester-etal-2021-power}. This focus on prompts receives validation from research, emphasizing its pivotal role in amplifying the reasoning competencies of LLMs \cite{kojima2023large}. 

There has also been a surge of research using LLMs for autonomous driving tasks. DriveLikeAHuman \cite{fu2023drive} and DiLu \cite{wen2023dilu} explore the potential of using an LLM to understand the driving environment in a human-like manner and analyze its ability to reason and interpret. DriveLM \cite{sima2023drivelm} is a dataset built upon nuScenes and CARLA, and proposes a VLM-based baseline approach for jointly performing GraphVQA and end-to-end driving.

\subsection{Critical Scenarios}
Ambiguous definitions of edge-case, boundary, and critical scenarios are commonplace in autonomous driving literature \cite{cai2022survey, nassif2023safety}. Edge-case scenarios typically refer to an infrequent or unique scenario that lies at the extreme limits within a distribution of scenarios (or environment specifications, i.e. scenario configurations omitting the trajectory of the ego-vehicle) \cite{drayson2023cc}. Evolutionary Algorithms (EAs) rank highly among the preferred methods for creating edge-case scenarios \cite{bevilacqua}. 

On the other hand, boundary scenarios are often defined as those scenarios whose execution results are in the proximate area around the boundary between safe and unsafe, i.e. perturbing the scenario parameters might lead to significantly different execution results \cite{zhang2021finding}. To generate these scenarios, the following methods are popular: Accelerated Evaluation \cite{Zhao_2017}, Genetic Algorithms (GA) \cite{li}, Particle Swarm Optimization \cite{klischat}, and so on \cite{machines10111101}. 

Critical scenarios are defined in the literature to be the union of edge-case scenarios and boundary scenarios \cite{vater2023systematic}, illustrated in Fig. \ref{fig:Scenario}. For clarity, and in hopes of easing the acquisition of unified safety standards by AV regulatory frameworks \cite{nassif2023safety}, we reiterate this distinction to partition scarcity and safety concepts that are too often bundled together for scenario categorization \cite{sun2021scenario}. 

Current research in AV training primarily focuses on predefined and dynamic scenarios. However, the use of LLMs for enhancing scenario generation, particularly for complex and critical situations, remains largely underexplored. There's a need for a comprehensive, open-source framework that generates critical scenarios to augment RL training data using LLMs and criticality metrics. This type of solution can adaptively create diverse and challenging scenarios, exponentiating AV safety and path planning development.

\section{METHODOLOGY}

\subsection{Reinforcement Learning for Autonomous Vehicle Environment}
An overview of our pipeline is shown in Fig. \ref{fig:arch}. We employed highway-env \cite{highway-env} as our base simulation environment, notable for its adjustable parameters like traffic dynamics and vehicle behaviors. Adjusting the parameters within the environmental configuration (such as vehicle density and behaviors) naturally changes the nature of scenarios exposed to the RL agent during training. 

Our AV path planning and prediction models utilize Proximal Policy Optimization (PPO), favored for striking an optimal balance between enhancing performance and maintaining training stability. This is primarily achieved through its unique objective function:

\begin{equation}
L^{CLIP}(\theta) = \hat{\mathbb{E}}_t \left[ \min(r_t(\theta) \hat{A}_t, \text{clip}(r_t(\theta), 1 - \epsilon, 1 + \epsilon) \hat{A}_t) \right]
\end{equation}

In this equation, $r_t(\theta)$ represents the ratio of the new to the old policy's probability, with $\hat{A}_t$ denoting the advantage estimate. The key aspect here is the $\epsilon$-clipped objective function that regulates the update size, ensuring gradual and stable learning improvements. For a more comprehensive insight into PPO and its significance within the context of deep reinforcement learning, refer to \cite{li2017deep}. The choice of PPO over other algorithms is guided by its proven efficacy in handling diverse and unpredictable scenarios, aligning with our goal of enhancing autonomous driving technologies.

\subsection{highD Dataset for Realistic Traffic Simulation}
\label{highd}
A critical component of our approach involves incorporating real-world traffic dynamics into our simulations. To achieve this, we incorporate the highD dataset \cite{highDdataset}, which offers a fairly comprehensive and realistic representation of vehicle behaviors observed at various times on highways. The highD dataset provides extensive recordings of natural vehicle trajectories on German highways (capturing details like vehicle type, size, maneuvers, precise positioning, etc.) and is collected through drone surveillance cameras to overcome common data collection limitations that may occur when gathering such data via conventional infrastructure sensors. Details of how we used the highD dataset, including our methodology for clustering driving behaviors and generating traffic scenarios, will be explained in Section \ref{sec:exp_highd}.

To diversify the training dataset, we need to edit the environment configuration after a certain number of training episodes (after a single epoch). To automate this process, we choose to leverage the highD dataset or via an LLM. This is largely a \textit{knowledge-driven} approach because we either exploit the real-world knowledge base of the highD dataset or the in-context knowledge and meaningful priors that are embedded within the LLM. In particular, we exploit knowledge relating to traffic objects and human driving behavior. This methodology fundamentally preserves natural language interpretability and augments the potential for specific scenario querying \cite{reward_design_with_LLMs}. 

\subsection{Risk Metrics to Measure Safety-Criticality}
In our framework, two primary surrogate safety measures are employed to evaluate the safety-criticality of maneuvers performed by the ego vehicle: the Time to Collision ($TTC$) and a Unified Risk Index ($r$).

$TTC$ is defined as:
\begin{equation}
    TTC = \frac{x_{\text{rel}}}{\lvert{v}\rvert_{\text{rel}}}
\end{equation}

Here, ${x_{\text{rel}}}$ denotes the relative distance between the two vehicles in question, and $\lvert{v}\rvert_{\text{rel}}$ implies the relative speed of the two vehicles. A lower TTC indicates a higher risk of collision. If $TTC$ at any timestep in a scenario drops below a predefined threshold value, we increment a $TTC$ \textit{near miss count}. 

The unified risk index ($r$) is a function of the longitudinal and lateral risk index ($r_{lon}$ and $r_{lat}$) \cite{candela2021quantitative, safe_and_efficient_leo}. To calculate $r_{lon}$ and $r_{lat}$, we must first ascertain the minimum safe longitudinal and lateral distances ($d_{\min}^{lon}$ and $d_{\min}^{lat}$). These are defined below:
\begin{equation}
d_{\min }^{l o n}=\left[v_r \rho+\frac{1}{2}\rho^{2}a_{\text{max}}+\frac{(v_r+\rho a_{max})^2}{2 b_{\text {min}}}-\frac{v_f^2}{2 b_{\text {max}}}\right]_{+}
\end{equation}

where $[x]_+:= \text{max}(x,0)$, a vehicle with velocity $v_r$ drives behind another vehicle (in the same direction) with velocity $v_f$. For any braking of at most $b_{\text{max}}$, the trailing vehicle has a response time of $\rho$ during which it accelerates by at most $a_{\text{max}}$, and immediately starts braking (after the response kicks in) by at least $b_{\text{min}}$. 

The minimum safe lateral distance ($d_{\min}^{lat}$) is given as the following: 
\begin{equation}
d_{\min}^{lat} = \left[\left(v^{ego}_{lat}\right)\rho + \frac{(v^{ego}_{lat})^2}{4 b_{\text{min}}} \right. 
 \left. - \left(\left(v^{nln}_{lat}\right)\rho + \frac{(v^{nln}_{lat})^2}{4b_{\text{min}}} \right) \right]_{+}
\end{equation}

With $v^{ego}$ and $v^{nln}$ being the velocity of the ego vehicle and the \textit{nearest lane neighbor} vehicle, $v_{lat}$ is the lateral speed of the specified vehicle, response time $\rho$, and braking of at least $b_{min}$. 

The longitudinal and lateral risk index, $r_{lon}$ and $r_{lat}$ respectively, is given by the conditional equation(s) below: 

\begin{equation}
    r_{lon}= 
\begin{cases}
1-\frac{d^{lon}}{d^{lon}_{min}}           & d^{lon}_{min} > d^{lon} \\
    0 & \text{otherwise} \\  
\end{cases}
\end{equation}

\begin{equation}
    r_{lat}= 
\begin{cases}
1-\frac{d^{lat}}{d^{lat}_{min}}           & d^{lat}_{min} > d^{lat} \\
    0 & \text{otherwise} \\  
\end{cases}
\end{equation}

The current relative longitudinal and lateral distances between the ego vehicle and another background traffic vehicle (usually the nearest neighbor vehicle) are denoted as $d^{lon}$ and $d^{lat}$. 

Multiplying the lateral and longitudinal risk indices, and adding risk propensity parameters $\beta > 0$ and $\gamma > 0$, the unified risk index ($r$) is defined as: 
\begin{equation}
r = \left(r_{lon}\right)^\beta\left(r_{lat}\right)^\gamma 
\end{equation}

In our study, we set both $\beta$ and $\gamma$ to $1$ for simplicity. This approach allows us to quantify risk in a way that comprehensively accounts for both longitudinal and lateral dynamics, crucial for assessing the safety of AV maneuvers. When this exceeds a predefined threshold value, we increment a \textit{unified risk index $r$ threshold count}. 

In our training framework, we evaluate environment configurations based on their criticality, determined by metrics such as TTC and $r$. After a specified number of epochs, we analyze the distribution of these configurations. Configurations that repeatedly surpass predefined thresholds in TTC and $r$ are marked for their propensity to create boundary scenarios, indicating their critical nature. Conversely, configurations with high criticality but low occurrence are categorized as edge-case scenarios. This dual categorization helps identify configurations that effectively represent critical scenarios. These identified configurations, both low in occurrence and high in criticality, are then used directly or fed into an LLM to inspire similar critical scenario generation. This process enriches the training by continuously presenting the RL agent with safety-critical challenges, effectively testing and enhancing its capabilities in handling real-world complexities.

\section{EXPERIMENTS AND RESULTS}
\subsection{Transforming highD Data for Scenario Generation}
\label{sec:exp_highd}
In our experiment, we categorized driving behaviors from the highD dataset into three distinct types: aggressive, defensive, and normal \cite{8492700}. To simulate these behaviors realistically, we developed specialized vehicle models with unique parameters like politeness and preferred following distance.

We employed the KPrototypes clustering method to analyze driving styles, focusing on variables such as speed, acceleration, and lane-changing behavior. This approach effectively grouped vehicles into behavior categories, providing essential insights for regenerating traffic scenarios in the HighwayEnv simulation environment. This process aims to replicate real-world traffic conditions for a thorough evaluation of autonomous driving algorithms under various realistic scenarios.

Furthermore, we emphasized the recreation of high-risk scenarios from the highD dataset. For this, we computed the Risk Perception (RP) metric at each timestep for every vehicle pair. The RP, calculated as $RP = \frac{A}{THW} + \frac{B}{TTC}$, where A is set to 1 and B to 4 \cite{kruber2019highway}, quantifies the risk level based on Time Headway (THW) and Time To Collision (TTC). By identifying instances with the highest RP values, we could accurately replicate the most critical and dangerous scenarios for our simulations.

Our analysis included 60 diverse traffic configurations from the highD dataset. We meticulously recorded details such as vehicle types, counts, and dynamic behaviors like speed and acceleration. We also included information on critical vehicle pairs. This comprehensive data was collated into a scenario database, summarized in Table \ref{tab:scenario_features}, to inform our simulation scenarios.

\begin{table}[ht]
\centering
\caption{Explanation of Scenario Features}
\label{tab:scenario_features}
\begin{tabular}{|l|p{6cm}|}
\hline
\textbf{Feature} & \textbf{Explanation} \\
\hline
num\_aggressive & Number of aggressive vehicles \\
\hline
num\_defensive & Number of defensive vehicles \\
\hline
num\_regular & Number of regular vehicles \\
\hline
num\_trucks & Total number of trucks \\
\hline
num\_cars & Total number of cars \\
\hline
density & Vehicle density \\
\hline
scenario & Unique scenario identifier \\
\hline
vehicle\_i & Location, speed, acceleration, and lane of Vehicle I \\
\hline
vehicle\_j & Location, speed, acceleration, and lane of Vehicle J \\
\hline
\end{tabular}
\end{table}

\subsection{Prompting the LLM}
Our pipeline, CRITICAL, utilizes LangChain \cite{LangChain}, a tool designed for leveraging language models, to enhance scenario analysis and generation in AV simulations. Integrating a diverse array of data and constraints, including real-world traffic patterns from the highD dataset, $TTC$ \textit{near miss counts}, and \textit{unified risk index $r$ threshold counts}, CRITICAL ensures that configuration modification suggestions are well-informed and realistic. This setup allows CRITICAL to use its embedded language models effectively for analyzing our data and configuration, thereby generating relevant and practical changes to our simulation scenarios.

For each simulation run, we systematically recorded configurations, their outcomes, and vehicle failure types, creating a rich history that aids in understanding trends and patterns. Based on insights from this historical data and real-world traffic behaviors, the language model suggests modifications to key simulation parameters.

Our approach involves creating a JSON dictionary for each scenario parameter, as detailed in Table \ref{tab:scenario_features}. The values within this dictionary are constrained within realistic ranges to reflect real-world driving behaviors and traffic conditions accurately. By employing this method, we ensure that our simulation scenarios remain authentic and grounded in reality. This structured approach, using a JSON format with predefined value limits for each parameter, facilitates efficient parsing of LLM outputs and allows for a flexible yet controlled adaptation of the simulation environment. 

\subsection{Computational Setup}
Our experiments were conducted on a system powered by an Intel(R) Xeon(R) w5-2455X processor and an NVIDIA A6000 graphics card, supported by 256GB of RAM.

\begin{figure}[th]
\centering
\includegraphics[width=.48\textwidth]{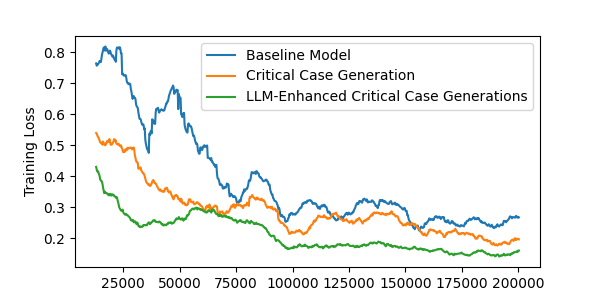}
\caption{Training loss comparison for Vanilla PPO (\textbf{Blue}), PPO with Critical Case Generation (\textbf{Orange}), and PPO with Critical Case Generation and Large Language Model Analysis (\textbf{Green}).}
\label{fig:loss}
\end{figure}

\subsection{Evaluation and Results}
We conducted evaluations comparing three distinct approaches to assess the effectiveness of our proposed CRITICAL framework. The first approach, the Baseline Model (PPO Only), utilizes the PPO algorithm in its original form. The second approach, Critical Case Generation, adopts our framework and characterizes it with an increased rate of critical scenario generation during training. The third and most comprehensive approach, LLM-Enhanced Critical Case Generation, combines the framework with LLM analysis, specifically employing the Mistral-7B-Instruct model \cite{jiang2023mistral}. This model was chosen for its exceptional capability in understanding and generating relevant, instructional content, vital for creating varied and intricate training scenarios.

All models were trained using a set of 50 different configurations selected from our scenario database and were evaluated on a separate set of 10 configurations. Each model was run ten times on each test configuration, and the average performance was calculated to determine the final results. The training loss curves shown in Fig. \ref{fig:loss} reveal the impact of the different training enhancements on model performance. The inclusion of critical scenarios indicates a more significant reduction in loss compared to the baseline model, suggesting an accelerated learning process. The LLM-Enhanced approach, which combines critical case generation and LLM analysis, exhibits the lowest and most stable loss, indicating superior learning efficacy and the potential for higher predictive accuracy in complex scenarios. 

The evaluation of our autonomous vehicle (AV) training models provided meaningful insights, as summarized in Table \ref{table:test}. The baseline model, utilizing just the PPO algorithm, recorded a reward of 50.068, an average episode length of 55.760, and a total of 89 crashes. The introduction of critical case generation without LLM led to enhanced performance, with the reward increasing to 61.527, the average episode length extending to 63.319, and crash counts dropping to 81.
\begin{table}[H]
\hspace{-3mm}
\caption{Testing results averaged over 10 runs.}
\begin{tabular}{|l|ccc|}
\hline
                                                                        & Reward $\uparrow$ & Episode Len $\uparrow$ & Crashes $\downarrow$\\ \hline
Baseline Model                                                               &  50.068   &    55.760      &  89    \\ \hline
Critical Case Generation                                                     &   61.527  &       63.319     &    81   \\ \hline
\begin{tabular}[c]{@{}l@{}}LLM-Enhanced Critical \\ Case Generations\end{tabular} &  \textbf{76.886}   &     \textbf{93.019}       &   \textbf{61}   \\ \hline
\end{tabular}
\label{table:test}
\end{table}
The most notable improvement was observed in LLM-enhanced critical case generation. This approach achieved a remarkable reward of 76.886, significantly increased the average episode length to 93.019, and reduced crash incidents to 61. These results not only highlight the enhanced navigational skills and safety capabilities of the model but also underscore the value of incorporating LLM analysis for crafting more complex and realistic training environments.

\begin{figure}[H]
\hspace{-5mm}
\centering
\includegraphics[width=.46\textwidth]{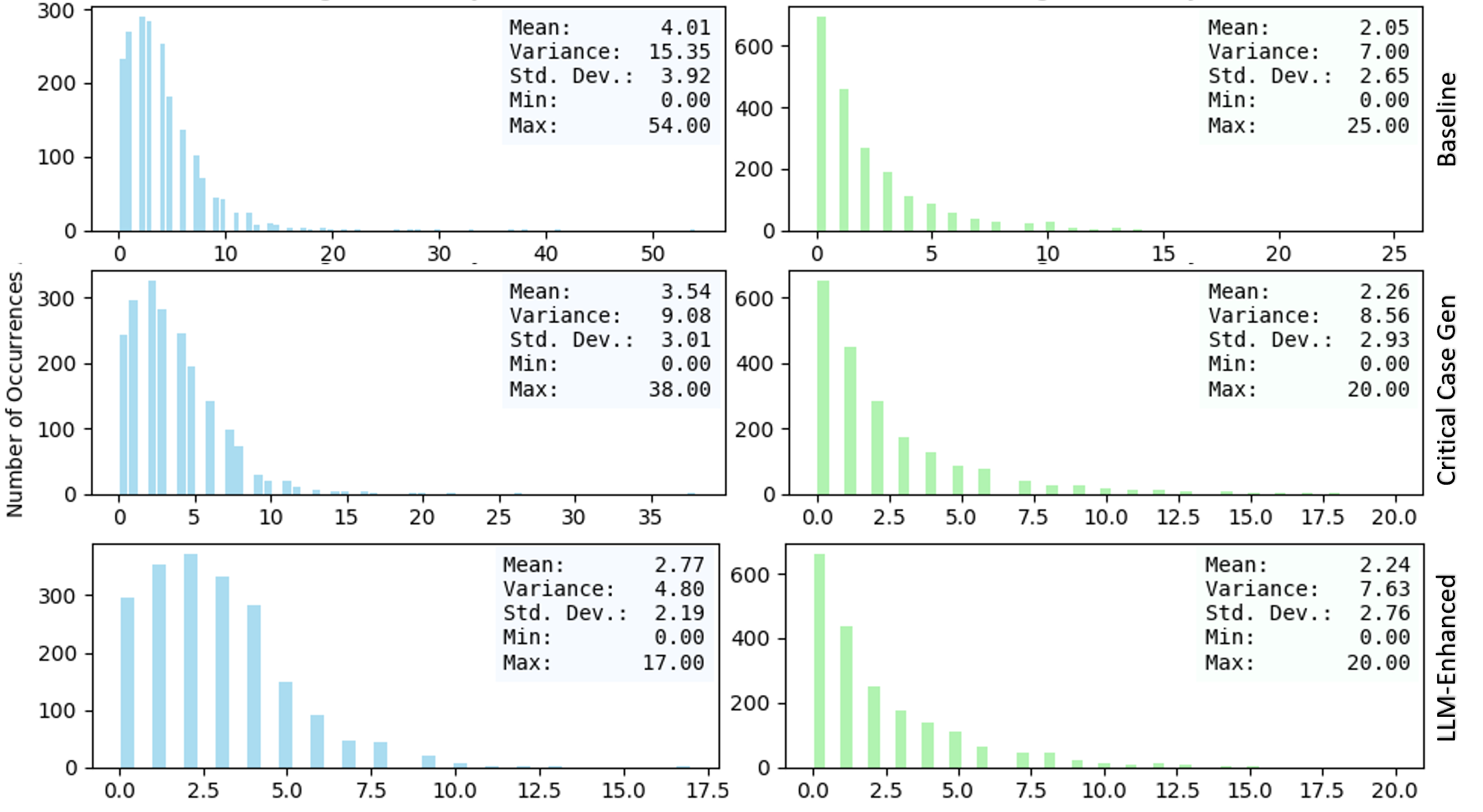}
\caption{The distributions on the right (\textbf{green}) are the configurations quantified by the $TTC$ \textit{near miss counts}, while the distributions on the left (\textbf{blue}) are the configurations by \textit{unified risk index $r$ threshold counts}. The distributions along the top row are the baseline. The middle row corresponds to Critical Case Generation without LLM, and the bottom row is LLM-Enhanced Critical Case Generations.}
\label{fig:risk_plot}
\end{figure}

\subsection{Critical Scenario Generation: Affects on RL-Agent Performance \& Criticality Measurements}
Figure \ref{fig:risk_plot} provides insights into the RL agent's performance. Throughout the training process, criticality scores and risk metrics, including \textit{unified risk index $r$} and \textit{TTC near-miss}, were continuously collected and analyzed. The baseline performance is illustrated in the first row. Criticality scores serve as indirect indicators of the agent's effectiveness, with low scores suggesting either competent performance or non-critical nature of the scenario. The second row shows how the agent fared when the framework actively sought out critical case generation without LLM. A noticeable reduction is observed in the criticality score associated with the \textit{$r$ threshold count}. The third row highlights LLM-enhanced configuration distributions, where a marked decrease in the mean value of \textit{$r$ threshold counts} strongly indicates superior agent performance across scenarios, translating to very low criticality scores. This demonstrates the framework's success in enriching the training dataset with critical scenarios to facilitate RL training, thereby confirming its effectiveness.

These findings underscore the efficacy of our approach, particularly the integration of LLM analysis in refining AV training. This method enriches the training process, preparing AVs for a wide range of real-world driving conditions and enhancing overall safety and performance.

\section{CONCLUSIONS and FUTURE WORK}

A novel framework (CRITICAL) is developed to support the training and testing of the RL-based AV control algorithm. This closed-loop framework incorporates real-world traffic dynamics from the highD dataset and surrogate safety measures to evaluate the criticality of scenarios encountered by a PPO agent during training. The closed-loop design of CRITICAL ensures that the RL agent is dynamically exposed to diverse challenging scenarios. This continual feedback creates a positive loop that enhances the agent's performance, improves the learning rate, and increases its robustness. 

The use of two surrogate safety measures (TTC and the unified risk index) in plotting configuration distributions provides insight into the evolving risk profile of scenarios encountered by the AV during training. Interestingly, we observe a decrease in the mean criticality of configurations as the AV's performance improves, reflecting its growing proficiency in handling challenging situations. This finding confirms the effectiveness of CRITICAL in improving the training of RL-based AV algorithms.

Future work will focus on expanding the validation of our framework across additional datasets beyond highD to ensure generalizability across different traffic conditions and road layouts. We also aim to test the efficacy of the framework with a wider variety of RL algorithms to assess its versatility and to optimize its performance across diverse training paradigms. Through this, we seek to establish CRITICAL as an open-source tool in the pursuit of safe and reliable autonomous vehicle navigation systems.

\bibliographystyle{IEEEtran}
\bibliography{IEEEabrv,main}

\end{document}